\documentclass[runningheads]{llncs}

\usepackage{eccv}

\usepackage{eccvabbrv}

\usepackage{graphicx}
\usepackage{booktabs}
\usepackage[table]{xcolor} 
\usepackage{multirow}
\usepackage{float}
\usepackage[ruled,vlined,linesnumbered]{algorithm2e}
\usepackage[accsupp]{axessibility}  
\usepackage{pifont}
\newcommand{\cmark}{\ding{51}} 
\newcommand{\xmark}{\ding{55}} 

\newcommand{\Ls}{\mathcal{L}_{\mathrm{S}}}
\newcommand{\Ldpo}{\mathcal{L}_{\mathrm{DPO}}}
\newcommand{\pitheta}{\pi_{\theta}}
\newcommand{\piref}{\pi_{\mathrm{ref}}}

\usepackage{hyperref}

\usepackage{orcidlink}

\begin{document}

\title{RSICCLLM: A Multimodal Large Language Model for Remote Sensing Image Change Captioning} 

\titlerunning{RSICCLLM}


\newcommand{\equalcontrib}{\textsuperscript{*}}
\newcommand{\corrauth}{\textsuperscript{\textdagger}}

\author{
Yelin Wang\inst{1,2,}\equalcontrib\orcidlink{0009-0003-8272-6066} \and
Zijia Song\inst{3,}\equalcontrib\orcidlink{0009-0008-5499-705X} \and
Shuo Ye\inst{2}\orcidlink{0000-0001-7756-8233} \and
Chuanguang Yang\inst{1}\orcidlink{0000-0001-5890-289X} \and
Miaoyu Wang\inst{1}\orcidlink{0009-0005-5086-3474} \and
Yong Xu\inst{3} \and
Zhulin An\inst{1,}\corrauth\orcidlink{0000-0002-7593-8293} \and
Yongjun Xu\inst{1}\orcidlink{0000-0001-6647-0986} \and
Zitong Yu\inst{2,4,}\corrauth\orcidlink{0000-0001-6505-3304}
}

\authorrunning{Y.~Wang et al.}



\institute{
State Key Laboratory of AI Safety, Institute of Computing Technology, Chinese Academy of Sciences, Beijing, China\\
\and
Great Bay University, Dongguan, China
\and
Harbin Institute of Technology, Shenzhen, China
\and
Dongguan Key Laboratory for Intelligence and Information Technology, Dongguan, China
}

\maketitle

\begingroup
\renewcommand{\thefootnote}{*}
\footnotetext{These authors contributed equally to this work.}
\renewcommand{\thefootnote}{\textdagger}
\footnotetext{Corresponding Authors.}
\endgroup

\begin{abstract}
Remote Sensing Image Change Captioning (RSICC) aims to describe changes between bi-temporal remote sensing images and holds significant research and application value. However, most existing methods rely on conventional deep learning architectures, and the limited model capacity constrains performance. Although large-model post-training techniques have achieved great success in general domains, their direct transfer to RSICC remains challenging due to data scarcity and the need for fine-grained change understanding.
To address this, we propose RSICCLLM, the first post-training framework for large vision-language models in RSICC. 
Specifically, we design\textbf{ a data generation paradigm}, release the instruction dataset RSICI, and establish \textbf{a task-specific RSICC benchmark}.
We further introduce \textbf{Difference-aware Supervised Fine-tuning} to explicitly extract change representations and guide the model in perceiving and understanding temporal differences. In addition, we propose \textbf{Dual-Negative Preference Optimization (DNPO)}, which employs two complementary negative-sample construction strategies to construct the preference dataset RSICP and further refine model performance. Extensive experiments validate the superior capability of RSICCLLM, which achieves outstanding results with only 7B parameters, surpassing models of substantially larger scales. The code and dataset will be made publicly available at \url{https://github.com/keaill/RSICCLLM}.

\end{abstract}

\vspace{-1.2em}
\section{Introduction}
 \vspace{-0.4em}
Remote sensing image change captioning (RSICC) generates descriptions of changes between bi-temporal remote sensing images \cite{zou2025rsicc_survey,hoxha2022changecaptioning,liu2022remote,ye20253mos,zhang2026multimodal,wang2026micro,zhu2026H-GAR}. Unlike remote sensing image change detection (RSICD), which reports only visual differences \cite{zhan2025skyeyegpt,liu2025pointsam,yu2026editorial,zhu2026delta,zhu2025emosym,zhu2025uniemo,he2026non}, RSICC maps visual evidence to semantics, improving interpretability and supporting many important public affairs, such as environmental assessment and disaster response. As an essential component of the remote sensing field, RSICC still faces multiple challenges \cite{zou2025rsicc_survey,wang2026affectagentcollaborativemultiagentreasoning,wang2026navigatingemotiontreehierarchical,yang2022cross,yang2024clip,huang2026preprompt,huang2024etag}, such as accurately perceiving real changes between bi-temporal images while ignoring irrelevant factors like illumination, and capturing subtle, diverse land-cover variations while aligning them with natural language. Therefore, RSICC holds substantial scientific and practical significance.

\begin{figure}[t]
  \centering
  \includegraphics[width=1\columnwidth]{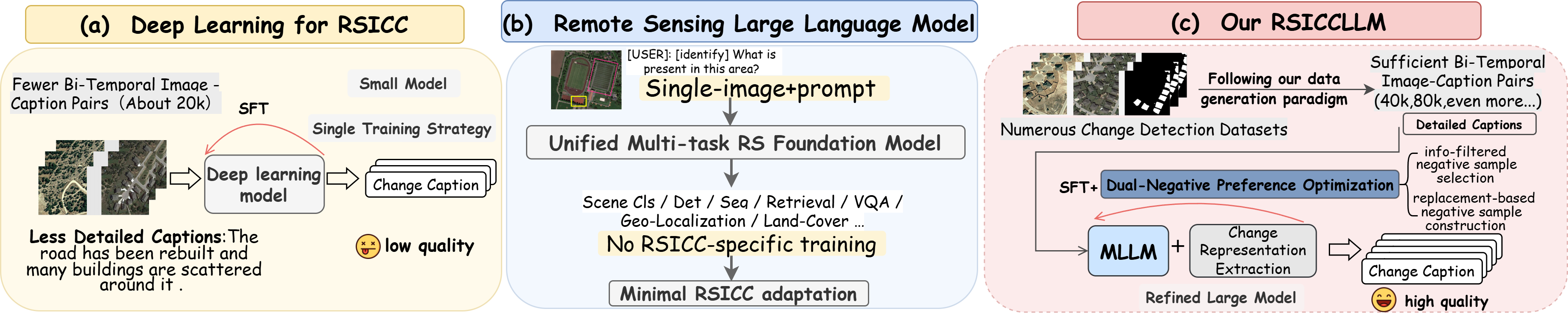}
  \vspace{-1.8em}
  \caption{The differences between the previous approaches and ours. (a) Previous methods are constrained by \textbf{limited model capacity} and \textbf{data scarcity}, hindering performance in change captioning. (b) Existing remote-sensing foundation models are rarely post-trained for RSICC with instruction or preference alignment, mainly due to limited RSICC data and preference supervision. (c) We introduce a data generation paradigm and train a refined larger model with Supervised Fine-tuning and Dual-Negative Preference Optimization, achieving superior results.}
  \label{fig:intro}
  \vspace{-2em}
\end{figure}

Recently, RSICC has attracted increasing research attention and made continuous progress. 
Semantic-CC \cite{Zhu2024SemanticCC} uses a SAM-based dual-image encoder with a lightweight multi-scale change detection (CD) decoder for pixel-level semantic guidance. Pix4Cap \cite{Liu2023Pix4Cap} adds a CD branch to produce pseudo-labels and a semantic-fusion augmentation to inject CD features into the captioner. 
Prompt-CC \cite{promptcc} introduces an image-level change classifier and a multi-prompt strategy that leverages an off-the-shelf caption generator to produce plausible captions without retraining.
However, these methods are primarily based on conventional deep learning frameworks, and their relatively small model capacity imposes an upper limit on overall performance.

Meanwhile, large-scale models and post-training techniques have shown strong success in general domains~\cite{deepseek-r1,qwen3} and are being introduced into remote sensing~\cite{geochat,xlrs}. For example, RS-GPT and Falcon leverage large-scale multimodal training to achieve strong performance on remote-sensing captioning and various downstream tasks~\cite{rsgpt,falcon}. Compared with conventional deep learning models, large models offer significant gains from increased capacity. However, due to the scarcity of specific datasets for RSICC and the limited ability of existing vision-language models to understand temporal changes, generating accurate and reliable change descriptions remains challenging. 

Motivated by these observations, we explore whether large-scale models and post-training techniques can be introduced into the RSICC field to enhance understanding of changes and achieve performance breakthroughs. Based on this idea, we propose RSICCLLM, the first\textbf{ post-training framework} of large models tailored for RSICC. The differences between the previous approaches and ours are shown in Fig. \ref{fig:intro}. 
We first construct \textbf{an instruction dataset generation paradigm} for RSICC using large-scale RSICD data. Specifically, we employ Qwen-VL-Max to generate instruction samples by feeding bi-temporal remote sensing images, together with their corresponding binary masks as geometric priors. By prompting the model to describe changes within the masked regions, we obtain a high-quality instruction dataset named \textbf{Remote Sensing Image Change Instruction dataset (RSICI)}. Based on this dataset, we establish \textbf{a task-specific RSICC benchmark}.
We then adopt Qwen2.5-VL-7B as the backbone for subsequent training. To further improve the model’s perception and understanding of temporal changes while suppressing irrelevant factors, we propose \textbf{Difference-aware Supervised Fine-tuning}, which explicitly extracts change representation and enhance the visual encoder with Central Difference Convolution (CDConv) \cite{cdc} and Hough Transform modules \cite{hough}. Using RSICI, we perform Supervised Fine-tuning (SFT) to obtain RSICCLLM-sft. 
Subsequently, we propose \textbf{Dual-Negative Preference Optimization (DNPO) }to further enhance model performance. We first design two strategies for constructing negative samples and build the \textbf{Remote Sensing Image Change Preference dataset (RSICP)}. In addition, we perform preference optimization on RSICCLLM-sft using RSICP and introduce a new loss operator to stabilize the training process, yielding the final model RSICCLLM.

Our main contributions can be summarized as follows:
\begin{itemize}
  \item We design \textbf{a new instruction dataset generation paradigm} in RSICC, building a large-scale instruction dataset RSICI and establish a task-specific \textbf{RSICC benchmark}. We also release the first preference dataset RSICP dedicated to this task.
  \item We propose RSICCLLM, the first post-training framework of large models for RSICC. To improve the model’s understanding of temporal changes, we propose \textbf{Difference-aware Supervised Fine-tuning.} And \textbf{a new preference optimization objective DNPO} is introduced to progressively refine the model’s performance.
  \item Extensive experiments demonstrate the superior performance of our approach. Notably, RSICCLLM achieves state-of-the-art performance across all evaluation metrics with only 7B parameters.
\end{itemize}
\vspace{-1em}
\section{Related Works}
\vspace{-0.8em}
\noindent\textbf{Remote Sensing Image Change Captioning.}
RSICC has recently emerged as a new research direction in remote sensing \cite{cheng2024change,wang2024advances,gu2024use,dong2024changeclip}. Early work by Chouaf \textit{et al}. \cite{chouaf2021captioning} explored RSICC on a private dataset with a VGG-16 encoder and an RNN decoder, which was later extended by Hoxha \textit{et al}. \cite{hoxha2022changecaptioning} using image- and feature-level fusion encoders. Liu \textit{et al}. \cite{liu2022remote} established the LEVIR-CC benchmark and proposed a CNN--transformer framework to emphasize changed regions. Subsequent studies further improved RSICC with progressive scale-aware modeling \cite{liu2023progressive} and Mamba-based spatio-temporal feature extraction \cite{chen2024rsmamba}.
All the aforementioned methods are deep-learning-based, and their progress is severely constrained by limited datasets and small model capacity. In contrast, our approach addresses these limitations by introducing a data generation paradigm and a large-scale training framework.

\noindent\textbf{Multimodal Reasoning.}\quad 
Recent work shifts multimodal reasoning from supervised learning to reinforcement-based fine-tuning with explicit reasoning \cite{wang2025perception,yang2025visionthink,shen2025semi,geng2025x}. FAST and TON reduce unnecessary computation via difficulty-aware control or a learned “think-or-not” policy \cite{fast_2025,ton_2025}. Chain-of-Focus improves region localization with RL-based adaptive search/zooming, and MiCo leverages cross-image contrastive learning with verifiable QA-free rewards \cite{chain_of_focus_2025,mico_2025}. UniVG\text{-}R1 advances visual referring through CoT initialization and GRPO-style optimization with reweighting \cite{univg_r1_2025}. Kimi\text{-}VL and G1 further highlight efficient MoE-style VLMs and RL gains after perception-first cold starts \cite{kimi_vl_2025,g1_2025}, while Qwen-AD advocates a look-before-you-decide strategy \cite{qwen_ad_2024}. These advances motivate a domain-specific RSICC framework for further breakthroughs.


\begin{figure*}[t]
\vspace{-0.9em}
  \centering
  \includegraphics[width=\textwidth]{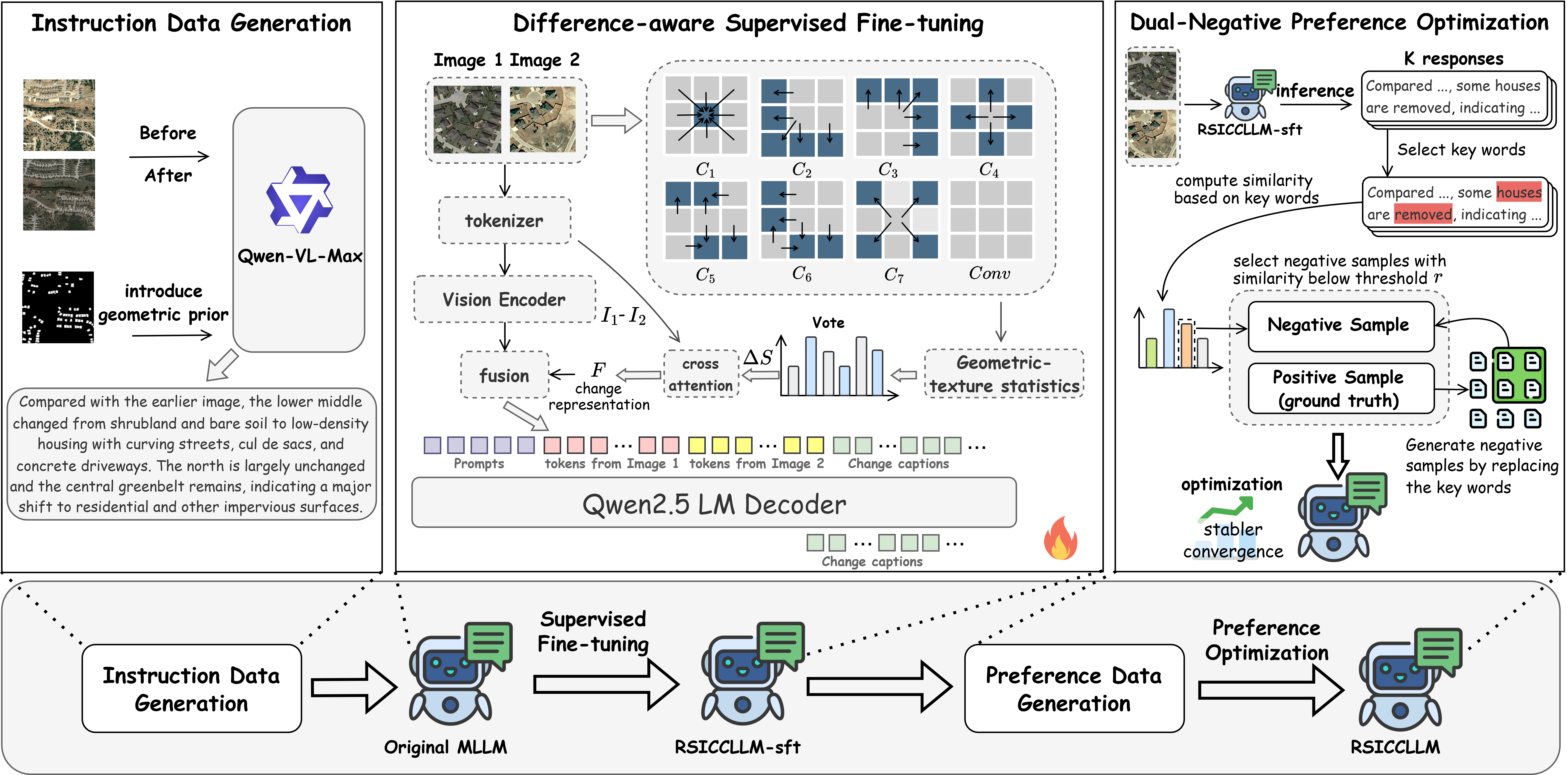} 
  \vspace{-1.8em}
  \caption{\textbf{Overview of the proposed RSICCLLM framework.}
  \textbf{Left}: Instruction Data Generation by Qwen-VL-Max \cite{Bai2023QwenVL} with the binary mask as geometric prior.
  \textbf{Middle}: Difference-aware Supervised Fine-tuning with Central Difference Convolution \cite{cdc}, asymmetric operators, and Soft-Hough voting \cite{hough} for explicitly extracting change representation. 
  \textbf{Right}: Dual-Negative Preference Optimization with two strategies to construct preference pairs, yielding the final RSICCLLM.}
  \label{fig:framework}
  \vspace{-0.6cm}
\end{figure*}

 \vspace{-0.8em}
\section{Methodology}
 \vspace{-0.4em}

\subsection{Overview of Framework}
\vspace{-0.2em}

We propose a three-stage RSICC framework, \textsc{RSICCLLM} (Fig.~\ref{fig:framework}): 
\textbf{1) Instruction Data Generation Stage.} We use Qwen-VL-Max to synthesize instruction samples from bi-temporal images, conditioned on a binary change mask as a geometric prior, with prompts eliciting explicit land-surface transition descriptions.
\textbf{2) Difference-aware Supervised Fine-tuning Stage.} We enhance the vision encoder by injecting manually extracted change representations into its forward pass, and then perform SFT with the generated instructions to obtain RSICCLLM-sft.
\textbf{3) Dual-Negative Preference Optimization Stage.} To enable preference training, we construct preference datasets using two complementary strategies, which are info-filtered negative sample selection and replacement-based negative sample construction. The resulting datasets are merged and used to further fine-tune RSICCLLM-sft via preference optimization, yielding RSICCLLM with stronger discriminative and descriptive abilities.

 \vspace{-0.8em}
\subsection{Instruction Data Generation}
\vspace{-0.2em}


In the data generation stage, 
we integrate the LEVIR-CD~\cite{chen2020spatial} and SYSU-CD~\cite{shi2021sysucd} datasets to automatically generate approximately 35k samples, among which 32k are used for training and 3k as \textit{\textbf{in-domain evaluation set}}, covering building changes as well as diverse man-made and natural changes. Meanwhile, we further integrate S2Looking~\cite{shen2021s2looking} and CDD~\cite{CDD} to automatically generate approximately 5k samples as \textit{\textbf{out-of-domain test set}}, which mainly consist of rural scenes, off-nadir building changes, and seasonally varying imagery, providing a more robust assessment for out-of-domain performance.
Each sample consists of a pre-change image, a post-change image, and a fine-grained textual description of the differences. 
For the Remote Sensing Image Change Instruction (RSICI) dataset, the instruction data generation process is shown in Algorithm~\ref{alg:instruct}. The data construction details and ablations are in \textbf{Appendix A}.

\begin{figure}[t]
\centering
\begin{minipage}{1\textwidth}
\begin{algorithm}[H]

\caption{Construction of \textsc{RSICC-Instruction} from LEVIR\_CD, SYSU\_CD, CDD, S2Looking}
\label{alg:instruct}
\KwIn{Image pairs $D$ (before, after) with binary masks $M$; Multimodal LLM; retry limit $R{=}2$}
\KwOut{Curated instruction dataset $\mathcal{D}$}

$\mathcal{D} \leftarrow \emptyset$ \tcp*{Initialize dataset}
\ForEach{$(I^{(1)}, I^{(2)}, M, \mathrm{meta}) \in D$}{
  Construct prompt $P$ using $(I^{(1)}, I^{(2)}, M)$ with geometric constraints and few-shot examples\;
  \For{$r=1$ \KwTo $R$}{
    Generate caption $y$ using the MLLM with prompt $P$\;
    Extract structured elements $\{object, direction, verb, quantity\}$ from $y$\;
    \If{change detected in mask $M$}{
      Check whether \textbf{object} and \textbf{verb} are correctly mentioned in $y$\;
    }
    \If{$y$ mentions words like ``mask'' or ``segmentation''}{
      Reject the sample and retry\;
    }
    \If{$y$ is excessively long}{
      Reject the sample and retry\;
    }
    \If{semantic hallucinations or off-topic content are detected}{
      Reject the sample and retry\;
    }
    \If{all conditions are satisfied}{
      Add $(I^{(1)}, I^{(2)}, M, y, \text{domain})$ to $\mathcal{D}$\;
      \textbf{break}
    }
    Adjust decoding strategy: change prompt length or examples\;
  }
}
Remove samples with undesirable keywords or formatting issues\;
Three domain experts re-check \emph{all} samples in $\mathcal{D}$; remove any sample that fails the quality criteria\;
Apply synonym substitution on extracted elements to augment linguistic diversity\;
\Return $\mathcal{D}$\;
\end{algorithm}
\end{minipage}
\vspace{-2.6em}
\end{figure}

\vspace{-0.6em}
\subsection{Difference-aware Supervised Fine-tuning}
\label{sec:Difference-aware supervised fine-tuning}



We perform Supervised Fine-tuning (SFT) on Qwen2.5-VL-7B for  change captioning, jointly modeling temporal dynamics and time-invariant spatial geometry to improve sensitivity to \emph{what} changed and \emph{where} it changed.

Given a bi-temporal image pair \(\text{Image}_{1}\) and \(\text{Image}_{2}\), we first perform differentiable spatial registration to remove non-semantic pixel shifts arising from sensor pose or terrain relief. Afterward, we tokenize the two images and compute a difference map as the initial change cue:
\begin{equation}
T \;=\;  I_1 - I_2  \;,
\end{equation}
where \(T\) encodes the magnitude of local intensity changes and highlights candidate change locations.


To extract time-invariant spatial geometry, we introduce eight geometry-sensitive convolution kernels inspired by Central Difference Convolution (CDConv) \cite{cdc}. Given bi-temporal remote-sensing images as input, these kernels can extract filtered geometric textures that are insensitive to illumination and weather. Specifically, they include one standard convolution operator \(C_{8}\); one CDConv operator \(C_{1}\); four directional asymmetric operators-\(C_{2}\), \(C_{3}\), \(C_{5}\), and \(C_{6}\); and two symmetric operators-\(C_{4}\) and \(C_{7}\). Their formats are shown in Fig.~\ref{fig:framework} and definitions are as follows:
\begin{equation}
\begin{split}
C^{t}_i(r_x,r_y) 
=& \sum_{(\Delta r_x,\Delta r_y)\in \mathcal{R}} \text{Image}_{t}(
   r_x + \Delta r_x ,r_y + \Delta r_y) \cdot\\ &(\sum_{(\Delta r_x,\Delta r_y)\in \mathcal{R}} \delta_i(\Delta r_x,\Delta r_y)\omega(\Delta r_x,\Delta r_y)),
\end{split}
\end{equation}
where $\mathcal{R}=\{(-1,-1),(-1,0),...,(1,0),(1,1)\}$ denotes the local receptive field of the trainable $3\times3$ vanilla convolution kernel $\omega$, and \(\text{Image}_{t}\in\mathbb{R}^{C\times H\times W}\) is the input feature map. $t \in \{1,2\}$ indicates the time index of images, and $i$ is the index of operators. \((r_x,r_y)\) represents the current pixel location, and $\delta_i(\Delta r_x,\Delta r_y)$ is a gate function to determine whether each convolution kernel element $\omega(\Delta r_x,\Delta r_y)$ participates in the computation.
We employ Hough Transform~\cite{hough}, and project each response map 
\(C_i^{t}\) (\(t\in\{1,2\}\)) into the Hough parameter space \((\rho,\theta)\)
to accumulate geometric evidence via voting:
\begin{equation}
  H_i^{t}(\rho,\theta)
  \;=\;
  \mathrm{HoughTransform}\!\big(C_i^{t}\big),
  t \in \{1,2\},
\end{equation}
where $i$ denotes the index of operators. After voting, we obtain the final bi-temporal texture statistic:
\begin{equation}
  H^{t} \;\triangleq\; \big\{\,H_i^{1}(\rho,\theta),\; H_i^{2}(\rho,\theta)\,\big\},
\end{equation}
where $\triangleq$ indicates the voting operation. Crucially, instead of differencing directly in the parameter space, we independently apply the inverse Hough transform to each time step to reconstruct the spatial structural responses:
\begin{equation}
  S^{(t)}
  \;=\;
  \mathrm{Hough}^{-1}\!\big(H^{t}(\rho,\theta)\big), t \in \{1,2\},
\end{equation}
where $S^{(t)}$ preserves complete geometric semantics for both bi-temporal images, avoiding the semantic loss caused by early differencing. We obtain a normalized geometric–texture features as follows:
\begin{equation}
\Delta S 
\;=\; 
\frac{S^{(2)} - S^{(1)}}{\lVert S^{(1)} \rVert_1 + \lVert S^{(2)} \rVert_1 + \epsilon}.
\end{equation}



To better represent the change patterns, we employ cross-attention to fuse the preliminary feature cues \(T\) with the geometric–texture features \(\Delta S\). Specifically, we treat \(T\) as the query vector and \(\Delta S\) as the key and value vectors. Through learnable projections, both the features \(T\) and \(\Delta S\) are mapped to the same dimension $d$, and the feature fusion process is computed as follows:
\begin{equation}
F\;=\; \mathrm{softmax}\!\left(\frac{T \Delta S^\top}{\sqrt{d}}\right) \Delta S \;\in\; \mathbb{R}^{N \times d},
\end{equation}
where $N$ denotes the number of image tokens after tokenization, and $F$ indicates the final change representation.
We then enhance the original image embeddings  \(V_{\mathrm{orig}} \in \mathbb{R}^{N \times d}\) from the vision encoder as follows:
\begin{equation}
  V_{\mathrm{enhanced}} \;=\; V_{\mathrm{orig}} + F,  
\end{equation}
where $V_{\mathrm{enhanced}}$ focuses more on the change regions within remote sensing images. The enhanced visual representations from the bi-temporal images are then combined with the prompt tokens and fed into the Qwen2.5 LM decoder to generate accurate change captions.

For SFT, we employ autoregressive (AR) objective as loss function. Given a ground truth $y=(y_1,\dots,y_T)$ corresponding to bi-temporal images, the formula is as follows:
\begin{equation}
\mathcal{L}_{\mathrm{AR}}
= -\sum_{t=1}^{T}\log P\!\left(y_t \mid y_{<t},\, I_1,\, I_2,\, P;\, \theta\right),
\end{equation}
where $I_1$ and $I_2$ denote the visual embeddings obtained from the two image inputs, and $P$ represents the prompt token. After
concatenating the visual embeddings and the prompt token, they are fed into the decoder with teacher forcing. $y_t$ is the $t$-th token in the ground-truth sequence, and $T$ denotes the sequence length. $\theta$ denotes the learnable parameters of the model. After Difference-aware Supervised Fine-tuning, we acquire the RSICCLLM-sft, which possesses preliminary perceptual and analytical capabilities, providing a solid foundation for further refinement through preference optimization.



\begin{figure*}[t!]
 \vspace{-1.3em}
  \centering
\includegraphics[width=0.95\textwidth]{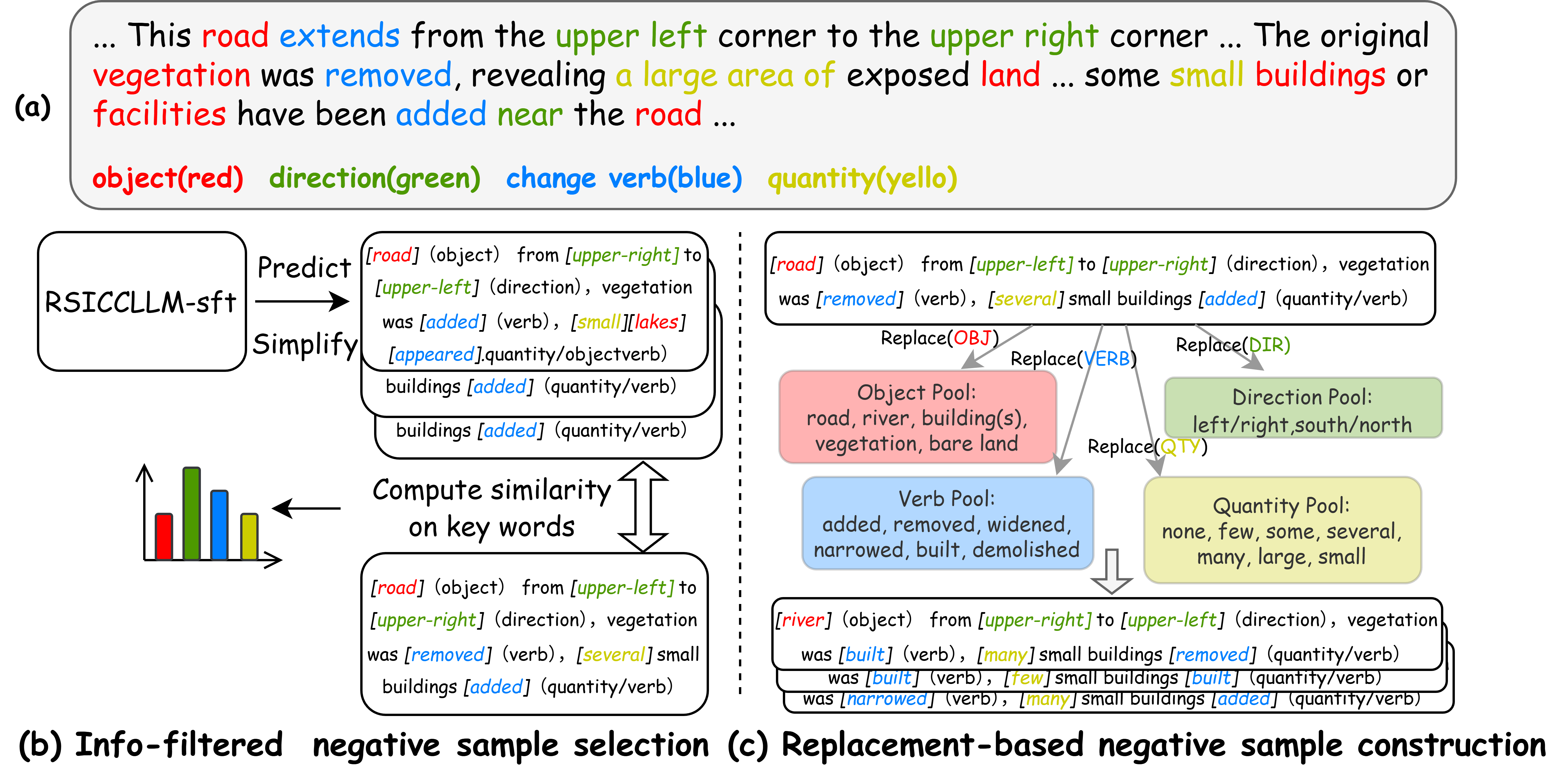}
\vspace{-1.4em}
  \caption{Two complementary negative sample construction strategies: (a) semantic decomposition of change captions into four key components; (b) info-filtered selection based on keyword similarity; and (c) replacement-based generation via controlled token substitution.}
  \label{fig:dpo}
\vspace{-2.4em}
\end{figure*}

 \vspace{-0.6em}
\subsection{Dual-Negative Preference Optimization}
Preference learning is a mainstream post-training technique that further improves generation quality beyond SFT by leveraging preference signals, even without explicit annotations. Its effectiveness, however, strongly depends on the quality of the preference dataset. For RSICC, the lack of explicit ground-truth labels makes quality assessment difficult, which in turn hinders the construction of high-quality preference data.
To address this issue, we propose Dual-Negative Preference Optimization, which constructs positive-negative sample pairs through two complementary strategies to facilitate preference learning.

We first summarize the characteristics of change captions, as illustrated in Figure \ref{fig:dpo}(a). Among these elements, object, direction, change verb, and quantity carry high information content and directly affect the sentence meaning; thus, they are defined as keywords. In contrast, articles, prepositions, and pronouns mainly serve to enrich sentence fluency and have limited impact on semantics, so they are referred to as non-keywords. The two proposed strategies for constructing the preference dataset are both designed based on this definition.


The first strategy is info-filtered negative sample selection and the process is shown in Fig.~\ref{fig:dpo}(b). For each input $x$ containing a pair of bi-temporal images, we use RSICCLLM-sft to generate $K$ candidate captions $\{y_k\}_{k=1}^{K}$. For any candidate caption $y=(w_1,\ldots,w_n)$, we define an indicator function $\phi(w_i)\in\{0,1\}$ to identify words with high information content.
Specifically, we set $\phi(w_i)=0$ for non-keywords, while assigning $\phi(w_i)=1$ to keywords with high information density. Accordingly, we obtain a filtered sequence composed only of the keywords as follows:
\begin{equation}
    y^{\mathrm{info}}=\big(w_i \in y\big|\phi(w_i)=1\big),
\end{equation}
where $y^{\mathrm{info}}$ denotes the filtered sequence for input $x$, which contains only the extracted keywords. The purpose of introducing $\phi(w_i)$ is to retain only the keywords that represent actual change facts, allowing for more accurate similarity computation. Similarly, for the ground-truth caption $y_{\mathrm{GT}}$, we obtain the corresponding filtered sequence $y_{\mathrm{GT}}^{\mathrm{info}}$. We then calculate the overall similarity between $y^{\mathrm{info}}$ and $y_{\mathrm{GT}}^{\mathrm{info}}$ using the averaged BLEU-1, BLEU-2, and ROUGE-1:
\begin{equation}
\begin{aligned}
S_{\mathrm{sim}}(y^{\mathrm{info}}, y_{\mathrm{GT}}^{\mathrm{info}})
&= \frac{1}{3}(\mathrm{BLEU}\text{-}1(y^{\mathrm{info}},y_{\mathrm{GT}}^{\mathrm{info}}) \\
&\quad+ \mathrm{BLEU}\text{-}2(y^{\mathrm{info}},y_{\mathrm{GT}}^{\mathrm{info}}) \\
&\quad+ \mathrm{ROUGE}\text{-}1(y^{\mathrm{info}},y_{\mathrm{GT}}^{\mathrm{info}})),
\end{aligned}
\end{equation}
where $S_{\mathrm{sim}}(y^{\mathrm{info}}, y_{\mathrm{GT}}^{\mathrm{info}})$ denotes the overall similarity between a candidate caption $y$ and the ground truth. For each candidate caption $y$ generated from the input $x$, we compute its similarity score following the above procedure and then select the negative sample set $\mathcal{N}_{\mathrm{sel}}$ for $x$ according to the following criterion:
\begin{equation}
\mathcal{N}_{\mathrm{sel}}
= \bigl\{y \in \{y_k\}_{k=1}^{K} \big|\tau_1 < S_{\mathrm{sim}}(y^{\mathrm{info}}, y_{\mathrm{GT}}^{\mathrm{info}}) < \tau_2 \bigr\},
\end{equation}
where $\tau_1$ and $\tau_2$ denote the lower and upper bounds of the similarity threshold, respectively. Through this constraint, we can identify an appropriate set of negative samples for each input, ensuring that they are neither too difficult nor too easy to distinguish from the positive examples.


The second strategy is replacement-based negative sample construction, which acquires negative samples by directly modifying the keywords in the ground truth $y_{\mathrm{GT}}$. And the process is shown in Fig.~\ref{fig:dpo}(c).  Specifically, we define the keyword set of the ground truth as $y_{\mathrm{GT}}^{\mathrm{info}} = \{ w \in y_{\mathrm{GT}} \mid \phi(w)=1 \}$ and introduce a replacement mapping $\psi: y_{\mathrm{GT}}^{\mathrm{info}} \to \mathcal{V}$, where $\mathcal{V}$ denotes a contrastive replacement pool built from the corpus. This pool is grouped by object, directions, change verbs, and quantity. When performing replacements, new words are selected only from the same group to preserve part-of-speech and morphological consistency, so that the resulting negative samples remain natural while being semantically incorrect in key aspects. For each ground-truth caption $y_{\mathrm{GT}}$, we randomly perform $M$ replacements to obtain the negative sample set $\mathcal{N}_{\mathrm{con}} = \{ \tilde{y}_m \}_{m=1}^{M}$. By combining the datasets from both approaches, we construct the final preference dataset containing 20k image pairs: 
\begin{equation}
\mathcal{D}_{\mathrm{pref}}
=\left\{\,\bigl(x,\ y_{w},\ y_{l}\bigr)\ \middle|\ y_{l}\in
\mathcal{N}_{\mathrm{sel}}\cup \mathcal{N}_{\mathrm{con}} \right\},
\end{equation}
where $y_w$ denotes the ground truth $y_{\mathrm{GT}}$. Through the synthesized preference dataset, we perform preference optimization with the following loss function:
\begin{equation}
  \mathcal{L}(\pi_\theta; \pi_{\text{ref}}) = \mathcal{L}_{\text{DPO}}(\pi_\theta; \pi_{\text{ref}}) + \lambda \mathcal{L}_{\text{S}}(\pi_\theta; \pi_{\text{ref}}), 
\end{equation}
where $\pi_\theta$ denotes the model continuously optimized during post-training, while $\pi_{\text{ref}}$ refers to RSICCLLM-sft. $\mathcal{L}_{\text{DPO}}(\cdot, \cdot)$ indicates the Direct Preference Optimization(DPO), and the formula is shown as follows:
\begin{equation}
\begin{aligned}
\mathcal{L}_{\text{DPO}}(\pi_\theta;& \pi_{\text{ref}}) 
= - \mathbb{E}_{(x,y_w,y_l) \sim \mathcal{D}_{\mathrm{pref}}} \left[ \log p(y_w \succ y_l \mid x) \right] \\
&= - \mathbb{E}_{(x,y_w,y_l) \sim \mathcal{D}_{\mathrm{pref}}} \Bigg[ 
\log \sigma \Bigg( 
\beta \log \frac{\pi_\theta(y_w \mid x)}{\pi_{\text{ref}}(y_w \mid x)} \\
&- \beta \log \frac{\pi_\theta(y_l \mid x)}{\pi_{\text{ref}}(y_l \mid x)} 
\Bigg) \Bigg],
\end{aligned}
\end{equation}
where minimizing $\mathcal{L}_{\text{DPO}}$ encourages $\pi_{\theta}$ to align with the best policy model, thereby increasing the likelihood of producing responses with higher quality. $\mathcal{L}_{\text{S}}(\cdot; \cdot)$, on the other hand, stabilizes training by constraining the divergence between the $\pi_\theta$ and $\pi_{ref}$. The formula is shown as follows:
\begin{equation}
\mathcal{L}_{\text{S}}(\pi_\theta;\pi_{\text{ref}}) = \mathbb{E}_{x\sim\mathcal{D}_{\mathrm{pref}}} \left[ D_{\mathrm{KL}}\left( \pi_\theta(\cdot \mid x) \,\|\, \pi_{\text{ref}}(\cdot \mid x) \right) \right].
\end{equation}
Here, we adopt the reverse KL divergence to prevent the policy model from deviating excessively from the reference model, thus ensuring stable training. 
Through our proposed DNPO, our model can leverage comprehensive preference information to achieve more stable preference training and generate higher-quality change captions.

 \vspace{-0.8em}
\section{Experiments}
 \vspace{-0.6em}
\begin{table*}[t]
\centering
\scriptsize
\setlength{\tabcolsep}{4pt}
\caption{Performance comparison of different models on our \textbf{\textit{in-domain}} RSICI test set to evaluate their capabilities for remote sensing image change captioning. The best results are highlighted in \textbf{bold}.}
\label{tab:contrast}
\vspace{-0.3cm}

\resizebox{\textwidth}{!}{%
\begin{tabular}{lcccccccc}
\toprule
\textbf{Method} & \textbf{BLEU-1} & \textbf{BLEU-2} & \textbf{BLEU-3} & \textbf{BLEU-4} & \textbf{ROUGE-1} & \textbf{ROUGE-2} & \textbf{ROUGE-L} & \textbf{SBS} \\
\midrule
Qwen-VL-Max \cite{Bai2023QwenVL}        & 27.55 & 12.12 & 5.18 & 2.70 & 34.20 & 6.86 & 20.10 & 71.79 \\
GLM-4.5V \cite{GLM2025GLM45V}           & 28.13 & 12.41 & 6.67 & 3.73 & 31.88 & 6.73 & 20.10 & 69.04 \\
SenseChat-Vision \cite{SenseChatVision2024} & 27.63 & 12.15 & 6.24 & 3.58 & 31.66 & 6.47 & 19.51 & 68.76 \\
Intern-S1 \cite{InternS12025TR}         & 26.47 & 11.97 & 5.47 & 3.00 & 31.98 & 6.78 & 19.18 & 70.89 \\
\midrule
InternVL-3.5-241B \cite{InternS12025TR} & 26.81 & 12.01 & 5.32 & 2.83 & 32.83 & 6.80 & 19.77 & 72.98 \\
Qwen3-VL-235B \cite{Qwen3VLRepo2025}     & 27.15 & 12.32 & 5.01 & 2.82 & 33.79 & 6.37 & 19.42 & 71.70 \\
Qwen3-VL-32B \cite{Qwen3VLRepo2025}      & 26.39 & 11.66 & 4.87 & 2.54 & 32.44 & 6.54 & 18.82 & 69.57 \\
Qwen2.5-VL-72B \cite{Bai2025Qwen25VL}    & 27.54 & 12.02 & 4.99 & 2.65 & 33.71 & 6.72 & 19.38 & 70.06 \\
\midrule
TEOChat-7B \cite{Irvin2024TEOChat}       & 8.63  & 2.74  & 1.37 & 0.94 & 18.89 & 3.18 & 11.13 & 40.97 \\
CCExpert-7B \cite{Wang2024CCExpert}      & 8.63  & 2.57  & 1.34 & 0.88 & 17.65 & 2.94 & 10.81 & 40.31 \\
\midrule
Qwen2.5-VL-7B \cite{Bai2025Qwen25VL}     & 7.40  & 2.35  & 1.08 & 0.61 & 17.27 & 2.79 & 10.16 & 42.06 \\
RSICCLLM-sft                              & 80.13 & 72.99 & 65.72 & 61.68 & 52.31 & 27.55 & 37.51 & 80.38 \\
\rowcolor{yellow!20}
RSICCLLM                                  & \textbf{81.28} & \textbf{74.70} & \textbf{66.81} & \textbf{62.78} & \textbf{53.54} & \textbf{28.94} & \textbf{38.30} & \textbf{82.72} \\
\bottomrule
\end{tabular}%
}
 \vspace{-1.4em}
\end{table*}

\subsection{Metrics and Implementation Details}
\textbf{Benchmark metrics. } We propose a new benchmark to evaluate model’s ability of analyzing changes between bi-temporal images, and employ BLEU-1, BLEU-2, BLEU-3, BLEU-4~\cite{papineni2002bleu}, ROUGE-1, ROUGE-2, and ROUGE-L~\cite{lin2004rouge} as evaluation metrics. BLEU-1 and ROUGE-1 measure the accuracy of key terms, BLEU-2 and ROUGE-2 reflect local semantic consistency, BLEU-3 and BLEU-4 assess sentence-level coherence, and ROUGE-L evaluates global semantic consistency. 
Given that change captions are open-ended and allow multiple valid phrasings, we introduce \emph{Sentence-BERT Similarity (SBS)} to evaluate the overall semantic similarity between the prediction and ground truth as follows:
\begin{equation}
\mathrm{SBS}=\frac{1}{N}\sum_{i=1}^{N}
\cos\!\big(\phi(\hat{y}_i),\,\phi(y_i)\big),
\end{equation}
where $N$ is the dataset size, and $\phi(\cdot)$ denotes a Sentence-BERT \cite{reimers2019sentence} encoder. $\hat{y}_i$ is the predicted caption, and $y_i$ denotes the reference. $\cos(\cdot, \cdot)$ represents the cosine similarity.
This metric can quantify the semantic gap between the outputs and the ground-truth captions. A higher value indicates better generation performance on RSICC.
We also design a metric to assess the hallucination robustness of our method against other models; details are provided in \textbf{Appendix~B.1}

\noindent\textbf{Implementation details. }All models are trained on $2$ NVIDIA A100 GPUs. More details are in \textbf{Appendix B.2}.

\begin{table*}[t]
\vspace{-1.0em}
\centering
\scriptsize
\setlength{\tabcolsep}{4pt}
\caption{Performance comparison of different models on our \textbf{\textit{out-of-domain}} RSICI test set for remote sensing image change captioning.}
\label{tab:out_contrast}
 \vspace{-0.85em}

\resizebox{\textwidth}{!}{%
\begin{tabular}{lcccccccc}
\toprule
\textbf{Method} & \textbf{BLEU-1} & \textbf{BLEU-2} & \textbf{BLEU-3} & \textbf{BLEU-4} & \textbf{ROUGE-1} & \textbf{ROUGE-2} & \textbf{ROUGE-L} & \textbf{SBS} \\
\midrule
Qwen-VL-Max \cite{Bai2023QwenVL} & 32.87 & 13.12 & 5.77 & 3.16 & 36.14 & 6.16 & 21.15 & 71.18 \\
GLM-4.5V \cite{GLM2025GLM45V} & 28.09 & 12.33 & 6.53 & 3.67 & 31.65 & 6.52 & 19.93 & 68.92 \\
SenseChat-Vision \cite{SenseChatVision2024} & 27.88 & 12.48 & 6.35 & 3.69 & 31.89 & 6.67 & 19.65 & 69.02 \\
Intern-S1 \cite{InternS12025TR} & 28.42 & 10.81 & 4.61 & 2.47 & 32.28 & 5.05 & 19.68 & 65.27 \\
\midrule
InternVL-3.5-241B \cite{InternS12025TR} & 28.67 & 11.08 & 4.66 & 2.47 & 32.58 & 5.22 & 19.70 & 65.45 \\
Qwen3-VL-235B \cite{Qwen3VLRepo2025} & 30.28 & 11.68 & 4.93 & 2.65 & 33.80 & 5.38 & 20.26 & 70.25 \\
Qwen3-VL-32B \cite{Qwen3VLRepo2025} & 29.33 & 10.95 & 4.63 & 2.49 & 33.88 & 5.19 & 20.35 & 64.69 \\
Qwen2.5-VL-72B \cite{Bai2025Qwen25VL} & 30.89 & 13.20 & 5.71 & 3.06 & 34.45 & 6.73 & 20.25 & 68.96 \\
\midrule
TEOChat-7B \cite{Irvin2024TEOChat} & 9.06 & 2.86 & 1.44 & 1.03 & 17.75 & 3.14 & 11.15 & 41.76 \\
CCExpert-7B \cite{Wang2024CCExpert} & 8.96 & 2.71 & 1.33 & 0.96 & 17.88 & 3.21 & 11.14 & 40.84 \\
\midrule
Qwen2.5-VL-7B \cite{Bai2025Qwen25VL} & 11.19 & 3.39 & 1.50 & 0.84 & 20.33 & 2.09 & 12.64 & 56.48 \\
\rowcolor{yellow!20}
RSICCLLM & \textbf{69.51} & \textbf{63.16} & \textbf{55.06} & \textbf{47.84} & \textbf{43.03} & \textbf{15.05} & \textbf{25.89} & \textbf{73.86} \\
\bottomrule
\end{tabular}%
}
 \vspace{-0.4em}
\end{table*}

\begin{table*}[t]
\vspace{-1em}
\centering
\scriptsize
\setlength{\tabcolsep}{4pt}
\caption{\textbf{Comparison with existing remote sensing text-image datasets.}
Avg\_L denotes the average number of words per caption.
Temporal indicates whether bi-temporal images are included in one sample.
Preference specifies whether the dataset contains preference information,
and ChangeCap denotes whether this dataset is designed for the RSICC task.}
 \vspace{-0.6em}
\label{tab:dataset_compare}

\resizebox{\textwidth}{!}{%
\begin{tabular}{l c c c c c c c c}
\toprule
\multirow{2}{*}{\textbf{Dataset}} &
\multirow{2}{*}{\textbf{Year}} &
\multirow{2}{*}{\textbf{\#Image}} &
\multicolumn{2}{c}{\textbf{Caption}} &
\multirow{2}{*}{\textbf{Temporal}} &
\multirow{2}{*}{\textbf{Preference}} &
\multirow{2}{*}{\textbf{ChangeCap}} &
\multirow{2}{*}{\textbf{Field}} \\
\cmidrule(lr){4-5}
& & & \textbf{\#Captions (Avg\_L)} & \textbf{Details} & & & & \\
\midrule

SpaceNet~7~\cite{van2021multi}            & 2021 & 2{,}389   & N/A           & \xmark & \cmark & \xmark & \xmark & Urban \\
S2Looking~\cite{shen2021s2looking}        & 2021 & 5{,}000   & N/A           & \xmark & \cmark & \xmark & \xmark & Urban \\
QFabric~\cite{verma2021qfabric}           & 2021 & 2{,}520   & N/A           & \xmark & \cmark & \xmark & \xmark & Urban \\
SpaceNet~8~\cite{hansch2022spacenet}      & 2022 & 2{,}576   & N/A           & \xmark & \cmark & \xmark & \xmark & Disaster \\
LEVIR-CC~\cite{liu2022remote}             & 2022 & 20{,}154  & 50{,}385 (40) & \xmark & \cmark & \xmark & \cmark & General \\
Dubai-CC~\cite{hoxha2022changecaptioning} & 2022 & 1{,}000   & 2{,}500 (35)  & \xmark & \cmark & \xmark & \cmark & Urban \\
RSICap~\cite{hu2023rsgpt}                 & 2023 & 2{,}585   & 2{,}585 (60)  & \cmark & \xmark & \xmark & \xmark & General \\
RS5M~\cite{zhang2024rs5m}                 & 2024 & 5M        & 5M (49)       & \cmark & \xmark & \xmark & \xmark & General \\
VRSBench~\cite{li2024vrsbench}            & 2024 & 29{,}614  & 29{,}614 (52) & \cmark & \xmark & \xmark & \xmark & General \\
WHU-CDC~\cite{shi2024multi}               & 2024 & 14{,}868  & 37{,}170 (--) & \cmark & \cmark & \xmark & \cmark & Disaster \\
XLRS-Bench~\cite{xlrs}            & 2025 & 934       & 934 (379)     & \cmark & \xmark & \xmark & \xmark & General \\
RSCC~\cite{chen2025rscc}                  & 2025 & 124{,}702 & 62{,}351 (72) & \cmark & \cmark & \xmark & \cmark & Disaster \\

\midrule
\rowcolor{yellow!20}
\textbf{RSICI (Ours)} & 2026 & 40{,}000 & 20{,}000 (77) & \cmark & \cmark & \xmark & \cmark & General \\
\rowcolor{yellow!20}
\textbf{RSICP (Ours)} & 2026& 20{,}000 & 10{,}000 (81) & \cmark & \cmark & \cmark & \cmark & General \\
\bottomrule
\end{tabular}%
}

\vspace{-0.4cm}
\end{table*}

 \vspace{-0.9em}
\subsection{Comparative Evaluation}
 \vspace{-0.4em}
We conduct evaluations on the in-domain test set of RSICI, with comparisons to mainstream multimodal models shown in Table~\ref{tab:contrast}. The compared methods include both large-scale general-purpose models and specialized models designed for change captioning. Clearly, our method, with only $7$B parameters, achieves the best performance across all $8$ metrics, with relative improvements of approximately 10\%–60\%. Overall, the results demonstrate that the proposed approach significantly surpasses both larger general-purpose and domain-specific models in terms of descriptive accuracy and semantic consistency. 


To further evaluate the model’s generalization ability, we conduct experiments on the out-of-domain test set of RSICI, with the results presented in Table \ref{tab:out_contrast}. It can be observed that even when tested on out-of-domain data, RSICCLLM can still effectively analyze changes between images and surpasses other larger models across all evaluation metrics, demonstrating its strong generalization capability.
Meanwhile, we use GPT-5.2 Thinking as the judge to evaluate four aspects---object, direction, action verb, and change magnitude---and report the average score across these attributes. Our model achieves the best performance. Detailed results are provided in \textbf{Appendix~B.3}.    

At this stage, we also compare our constructed datasets with other existing datasets, and the results are shown in Table~\ref{tab:dataset_compare}. RSICI contains a total of 40,000 images paired with 20,000 captions, with an average caption length of 77 words. It is the largest instruction dataset for RSICC in the general domain, providing sufficient training data for models to learn how to understand and analyze changes in bi-temporal images. RSICP includes 20,000 images and represents the first released preference dataset in the RSICC field. This dataset enables the model to more precisely distinguish changes between remote sensing images and plays an important role in advancing research in this area. 

\begin{figure}[t]
  \centering
  \includegraphics[width=0.85\columnwidth]{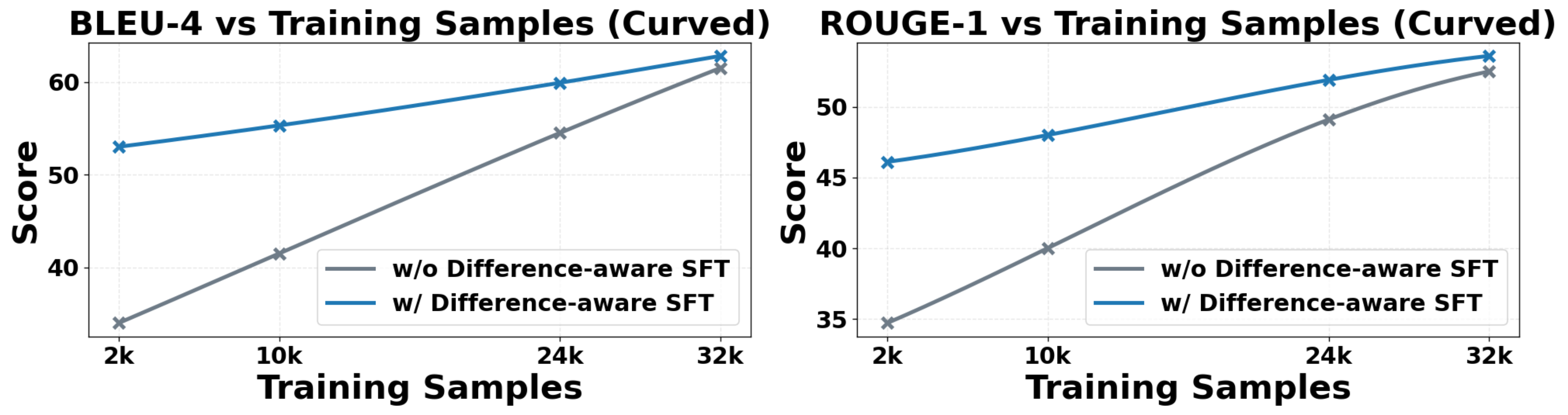}
   \vspace{-0.6em}
  \caption{\textbf{Ablation on Difference-aware SFT with different training sizes.} Each subfigure plots the metric score (vertical axis) against training size (horizontal axis).}
  \label{fig:few_samples}
\end{figure}

\begin{table}[t]
\vspace{-0.8em}
\centering
\scriptsize
\setlength{\tabcolsep}{3pt}
\renewcommand{\arraystretch}{0.92}

\begin{minipage}[t]{0.49\linewidth}
\centering
\vspace{0pt}

\captionof{table}{\textbf{Ablation on Difference-aware SFT components.} G denotes the geometric-texture statistics.}
\label{tab:module_ablation}
\vspace{-1.2em}
\resizebox{\linewidth}{!}{
\begin{tabular}{lcccc}
\toprule
\textbf{Method} & \textbf{BLEU-4} & \textbf{ROUGE-1} & \textbf{ROUGE-2} & \textbf{ROUGE-L} \\
\midrule
w/o G         & 62.54 & 53.13 & 28.20 & 38.11 \\
w/o $\Delta S$& 61.90 & 52.89 & 28.14 & 37.85 \\
w/o $T$       & 60.45 & 51.37 & 26.98 & 36.03 \\
\rowcolor{yellow!20}
RSICCLLM      & \textbf{62.78} & \textbf{53.54} & \textbf{28.94} & \textbf{38.30} \\
\bottomrule
\end{tabular}}

\captionof{table}{\textbf{Effectiveness of training strategies on various metrics.}}
\label{tab:sft_dpo_effect}
\vspace{-1em}
\resizebox{\linewidth}{!}{
\begin{tabular}{lcccc}
\toprule
\textbf{Strategy} & \textbf{BLEU-4} & \textbf{ROUGE-1} & \textbf{ROUGE-2} & \textbf{ROUGE-L} \\
\midrule
Without SFT & 3.21  & 17.27 & 2.79  & 10.94 \\
SFT         & 61.68 & 52.31 & 27.55 & 37.51 \\
\rowcolor{yellow!20}
SFT + DNPO  & \textbf{62.78} & \textbf{53.54} & \textbf{28.94} & \textbf{38.30} \\
\bottomrule
\end{tabular}}

\end{minipage}
\hfill
\begin{minipage}[t]{0.49\linewidth}
\centering
\vspace{0pt}

\captionof{table}{\textbf{Ablation on two negative sample construction strategies under DNPO.} Here, S = SFT the model with RSICI; F = preference training with dataset only from info-filtered negative sample selection, R = preference training with dataset only from replacement-based negative sample construction.}
\label{tab:dnpo_ablation_short}
\vspace{-1.4em}
\resizebox{\linewidth}{!}{
\begin{tabular}{lcccc}
\toprule
\textbf{Setting} & \textbf{BLEU-4} & \textbf{ROUGE-1} & \textbf{ROUGE-2} & \textbf{ROUGE-L} \\
\midrule
S      & 61.68 & 52.31 & 27.55 & 37.51 \\
S+F    & 62.38 & 53.19 & 28.44 & 38.03 \\
S+R    & 62.16 & 53.37 & 28.19 & 37.92 \\
\rowcolor{yellow!20}
S+F+R  & \textbf{62.78} & \textbf{53.54} & \textbf{28.94} & \textbf{38.30} \\
\bottomrule
\end{tabular}}

\end{minipage}
\vspace{-1.8em}
\end{table}

 \vspace{-0.8em}
\subsection{Ablation Study}
 \vspace{-0.2em}

\textbf{Ablation on Difference-aware SFT with different training size.}
Under the same implementation, we compare the model trained with the proposed Difference-aware SFT against the standard SFT on training sets of $2$k, $10$k, $24$k, and $32$k, evaluated by BLEU-4 and ROUGE-1. As shown in Fig.~\ref{fig:few_samples}, both models benefit from more data, while the model with Difference-aware SFT consistently outperforms the baseline. Notably, the performance gain becomes more significant in low-data regimes, indicating Difference-aware SFT can effectively guide the model to understand changes in bi-temporal images, especially when training data are limited. More results are shown in \textbf{Appendix C.1}.

\noindent\textbf{Ablation on different components in Difference-aware SFT.}
We analyze the contribution of the three modules in Difference-aware SFT separately. \textit{w/o G} denotes replacing the geometric-texture statistics with simple averaging over convolutional outputs. The definitions of $\Delta S$ and $T$ follow Sec.~\ref{sec:Difference-aware supervised fine-tuning}, and the results are reported in Table~\ref{tab:module_ablation}. Removing $T$ leads to the largest performance drop, underscoring the importance of explicit change extraction. Removing $\Delta S$ also degrades performance, indicating that the convolutional operators can provide useful change cues. Eliminating G further reduces accuracy, showing the effectiveness of geometric-texture statistics. Overall, these results confirm the benefit of all three components for change perception and understanding.

\noindent\textbf{Effectiveness of different training strategies.}
We evaluate the contribution of each training stage in Table~\ref{tab:sft_dpo_effect}. The base model performs poorly without fine-tuning, while Difference-aware Supervised Fine-tuning (SFT) brings substantial gains across all metrics. Adding the DNPO stage yields further improvements by exploiting constraints from positive–negative sample pairs to refine the model’s outputs.
We therefore construct a more challenging subset by manually selecting approximately 400 difficult samples, and find that DNPO brings a 7.4\% improvement over the SFT stage. Detailed results are provided in \textbf{Appendix~C.2}.

\noindent\textbf{Ablation on two negative sample construction strategies under DNPO.}
To assess the impact of two negative sample construction strategies, we conduct an ablation as shown in Table~\ref{tab:dnpo_ablation_short}. Clearly, both strategies individually yield consistent gains over the SFT-only baseline, while combining their preference datasets achieves the best performance, confirming the effectiveness of both strategies for constructing high-quality preference data.

\noindent\textbf{Ablation on cross attention fusion}.
Cross-attention is used to fuse information from different views. $T$, obtained via the MLLM tokenizer, is better aligned with semantic information but lacks explicit spatial-geometric cues. In contrast, the convolution-derived $\Delta S$ focuses on spatial geometry but cannot capture semantics. Fusing $T$ and $\Delta S$ therefore strengthens change perception and understanding. We modified the corresponding module in RSICCLLM, trained it with SFT only, and evaluated on the in-domain test set. The results are reported in Table~\ref{tab:deltaS_inject}, where $T$ denotes $I_1-I_2$.

\begin{table*}[t]
\vspace{-0.9em}
\centering

\begin{minipage}[t]{0.4\textwidth}
\centering
\scriptsize
\fontsize{6.0}{6.8}\selectfont
\setlength{\tabcolsep}{1.8pt}
\renewcommand{\arraystretch}{1.12}

\caption{\textbf{Ablation on cross attention fusion and $\Delta S$ injection strategies.}}
\label{tab:deltaS_inject}
\vspace{-1.4em}
\begin{tabular}{l c c c}
\hline
\textbf{$\Delta S$ Injection} & \textbf{BLEU-1} & \textbf{ROUGE-1} & \textbf{SBS} \\
\hline
Only $\Delta S$ & 78.29 & 50.63 & 78.27 \\
Concat ($[T, \Delta S]$) & 78.63 & 50.81 & 78.67 \\
Add ($T + \Delta S$) & 79.14 & 51.07 & 79.38 \\
Mul ($T \odot \Delta S$) & 79.39 & 51.24 & 79.42 \\
\rowcolor{yellow!20}
Cross-Attn & \textbf{81.28} & \textbf{53.54} & \textbf{80.38} \\
\hline
\end{tabular}

\vspace{-16pt}
\end{minipage}
\hfill
\begin{minipage}[t]{0.55\textwidth}
\centering
\scriptsize
\fontsize{6.0}{7.8}\selectfont
\setlength{\tabcolsep}{1.8pt}
\renewcommand{\arraystretch}{1.12}

\caption{\textbf{Comparison of SBS, parameters and throughput. \emph{samples/s} denotes the number of samples processed per second.}}
\label{tab:perf_comp}
\vspace{-1.4em}
\begin{tabular}{p{0.38\columnwidth} c r c}
\hline
\textbf{Method} & \textbf{SBS} $\uparrow$& \textbf{Param.}$\downarrow$ & \textbf{samples/s}$\uparrow$ \\
\hline
Qwen3-VL-235B & 71.70 & 235B & 0.104 \\
Qwen3-VL-72B & 70.06 & 72B & 0.213 \\
InternVL-3.5-241B & 72.98 & 241B & 0.096 \\
\rowcolor{yellow!20}
\textbf{Ours} & \textbf{82.72} & \textbf{7B} & \textbf{0.631} \\
\hline
\end{tabular}

\end{minipage}
\vspace{-1em}
\end{table*}

\begin{figure*}[t]
    \centering
    \includegraphics[width=1\textwidth]{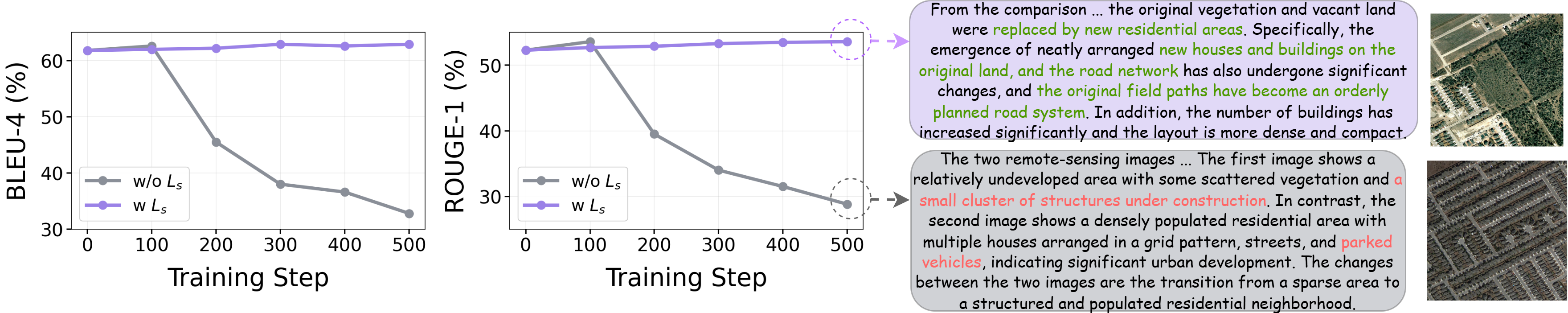}
   \vspace{-2.2em}
    \caption{Ablation study on whether to employ $\mathcal{L}_{\text{S}}$.}
    \label{fig:ablation_l_s}
    \vspace{-0.7cm}
\end{figure*}

\noindent\textbf{Ablation Study on \texorpdfstring{$\Ls$}{L_S}.}
\begin{figure}
 \vspace{-0.8em}
    \centering
    \includegraphics[width=1\linewidth]{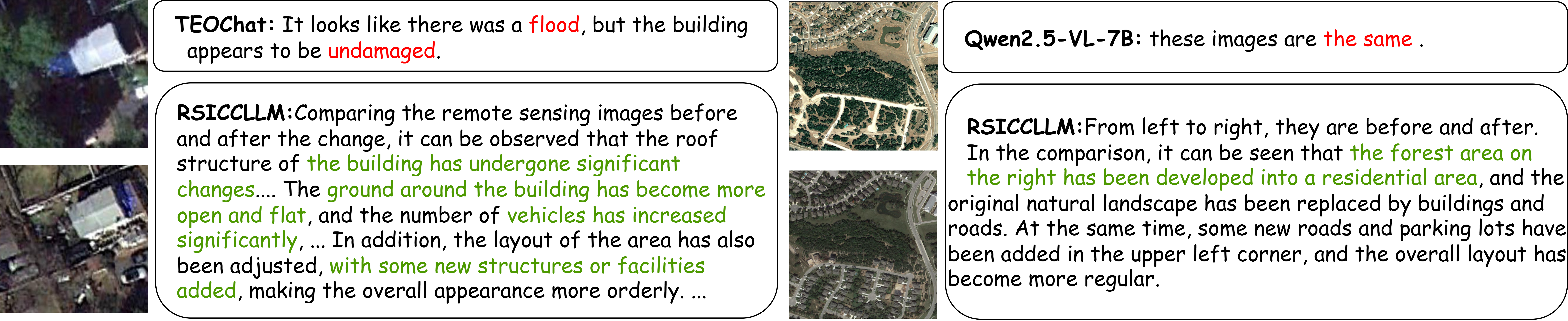}
    \vspace{-0.5cm}
    \caption{\textbf{Visual analysis of change captioning from different models for qualitative comparison.}}
    \label{fig:vis}
     \vspace{-1.8em}
\end{figure}
We conduct this ablation study to evaluate the impact of the loss term $\Ls$.
BLEU-4 and ROUGE-1 are adopted as evaluation metrics, and the results are shown
in Fig.~\ref{fig:ablation_l_s}. As observed, the model trained with the  $\Ls$
constraint performs slightly worse at the early stage compared to the one
trained only with $\Ldpo$. However, the $\Ldpo$-only model quickly reaches its
peak and then suffers a sharp performance drop as training continues, eventually
leading to collapse. In contrast, the model incorporating $\Ls$ exhibits a more
stable training process and ultimately achieves better and more consistent
performance through the joint effect of $\Ls$ and $\Ldpo$. These results suggest
that $\Ls$ helps prevent $\pitheta$ from deviating excessively from $\piref$,
mitigating the influence of data noise and improving the stability and
effectiveness of preference training.

\noindent\textbf{About inference cost and model comparison.}
As shown in Table~\ref{tab:perf_comp}, We compare our model with Qwen3-VL-235B, Qwen3-VL-72B, and InternVL-3.5-241B in terms of parameters, throughput, and performance. Despite using only a 7B model, we outperform the second-best model, Qwen3-VL-235B, by about 11 SBS points, while achieving a $\sim$6$\times$ higher prediction throughput in samples per second.
The experimental results compared with existing deep learning methods in the RSICC literature are provided in \textbf{Appendix~C.3}.

\vspace{-0.8em}
\subsection{Visual Analysis}
\vspace{-0.6em}

To further analyze the model’s behavior in the change captioning process, we visualize the outputs as shown in Fig.~\ref{fig:vis}. The upper example compares RSICCLLM with the domain-specific model TEOChat. Compared with TEOChat, our method better captures the changes between bi-temporal images. The lower example compares RSICCLLM with the baseline Qwen2.5-VL-7B, where RSICCLLM shows substantially improved change captioning ability, perceiving fine-grained changes, correctly localizing them, and producing more detailed descriptions. Additional visualizations are provided in \textbf{Appendix D}.

\vspace{-0.8em}
\section{Conclusion}
\vspace{-0.6em}
In this paper, we propose RSICCLLM, a domain-adapted post-training framework for vision-language models. By integrating Difference-aware Supervised Fine-tuning and Dual-Negative Preference Optimization, our method effectively aligns visual changes with high-quality descriptions. In addition, we design a new data generation strategy and release a large-scale instruction dataset RSICI. Meanwhile, we also establish the first preference dataset RSICP for this task. Extensive experiments demonstrate that RSICCLLM achieves superior performance across multiple metrics, validating the effectiveness of our approach.

\noindent\textbf{Acknowledgement.}
This work was supported in part by the National Natural Science Foundation of China under Grant Nos. 62576076, 62476264, and 62406312; the Guangdong Basic and Applied Basic Research Foundation under Grant No. 2023A1515140037; the CCF-Tencent Rhino-Bird Open Research Fund; the Guangdong Research Team for Communication and Sensing Integrated with Intelligent Computing under Project No. 2024KCXTD047; and the Youth Key Project of the Chinese Academy of Sciences under Grant No. GFQN-2026-34. The authors also acknowledge the SongShan Lake HPC Center (SSL-HPC) at Great Bay University for providing computational resources.

%
%
\bibliographystyle{splncs04}
\bibliography{main}
\end{document}